\documentclass[graybox]{svmult}

\usepackage{mathptmx}       
\usepackage{helvet}         
\usepackage{courier}        
\usepackage{type1cm}        
\usepackage{makeidx}         
\usepackage{graphicx}        
\usepackage{multicol}        
\usepackage[bottom]{footmisc}

\makeindex             

\begin{document}

\title*{Open science in machine learning}
\author{Joaquin Vanschoren and Mikio L. Braun and Cheng Soon Ong}
\institute{Joaquin Vanschoren \at Leiden University, Leiden, Netherlands, \email{joaquin@liacs.nl}
\and Mikio L. Braun \at TU Berlin, Berlin, Germany, \email{mikio.braun@tu-berlin.de}
\and Cheng Soon Ong \at National ICT Australia, Melbourne, Austrialia, \email{chengsoon.ong@unimelb.edu.au}
}
%
%
\maketitle

\abstract*{We present OpenML and mldata, open science platforms that provides easy access to machine learning data, software and results to encourage further study and application. They go beyond the more traditional repositories for data sets and software packages in that they allow researchers to also easily share the results they obtained in experiments and to compare their solutions with those of others.}

\abstract{We present OpenML and mldata, open science platforms that provides easy access to machine learning data, software and results to encourage further study and application. They go beyond the more traditional repositories for data sets and software packages in that they allow researchers to also easily share the results they obtained in experiments and to compare their solutions with those of others.}


\section{Introduction}

Research in machine learning and data mining can be speeded up tremendously by moving empirical research results ``out of people's heads and labs, onto the network and into tools that help us structure and alter the information''~\cite{Nielsen2008}. The massive streams of experiments that are being executed to benchmark new algorithms, test hypotheses or model new data sets have many more uses beyond their original intent, but are often discarded or their details are lost over time. In this paper, we present recently developed infrastructures that aim to make machine learning research more open. They go beyond the more traditional repositories\footnote{Well-known examples are the UCI repository, (\texttt{http://archive.ics.uci.edu/ml}), myExperiment (\texttt{http://myexperiment.org}) and MLOSS (\texttt{http://mloss.org}).} for data sets, implementations and workflows in that they allow researchers to also share detailed results obtained in experiments and to compare their solutions with those of others.

This \emph{collaborative} approach to experimentation allows researchers to share all code and results that are possibly of interest to others, which may boost their visibility, speed up further research and applications, and engender new collaborations. Indeed, many questions about machine learning algorithms can be answered on the fly by querying the combined results of thousands of studies on all available data sets. This facilitates much larger-scale machine learning studies, yielding more generalizable results~\cite{Hand2006}. Last but not least, these infrastructures keep track of experiment details, ensuring that we can easily reproduce them later on, and confidently build upon earlier work~\cite{Hirsh2008}.

\section{OpenML}
OpenML ({\footnotesize \texttt{http://openml.org}}) is a website where researchers can share their data sets, implementations and experiments in such a way that they can easily be found and reused by others. It offers a web API through which new resources and results can be submitted automatically, and is being integrated in a number of popular machine learning and data mining platforms, such as Weka, RapidMiner, KNIME, and data mining packages in R, so that new results can be submitted automatically. Vice versa, it enables researchers to easily search for certain results (e.g. evaluations of algorithms on a certain data set), to directly compare certain techniques against each other, and to combine all submitted data in advanced queries.

To make experiments from different researchers comparable, OpenML uses \emph{tasks}, well-described problems to be solved by a machine learning algorithm or workflow. A typical task would be: \emph{Predict (target) attribute X of data set Y with maximal predictive accuracy}. Similar to a data mining challenge, researchers are thus challenged to build algorithms or workflows that solve these tasks. Tasks can be searched online, and will be generated on demand for newly submitted data sets.

Tasks contain all necessary information to complete it, always including the input data and what results should be submitted to the server. Some tasks offers more structured input and output: predictive tasks, for instance, include train and test splits for cross-validation, and a submission format for all predictions. The server will evaluate the predictions and compute scores for various evaluation metrics.

An attempt to solve a task is called a \emph{run}, and includes the task itself, the algorithm or workflow (i.e., \emph{implementation}) used, and a file detailing the obtained results. These are all submitted to the server, where new implementations will be registered. For each implementation, an online overview page is generated summarising the results obtained over all tasks, over various parameter settings. For each data set, a similar page is created, containing a ranking of implementations that were run on tasks with that data set as input.

OpenML provides a REST API for downloading tasks and uploading data sets, implementations and results. This API is currently being integrated in various machine learning platforms such as Weka, R packages, RapidMiner and KNIME~\footnote{Beta versions of these integrations can be downloaded from the OpenML website.}.

To make the shared results maximally useful, OpenML links various bits of information together in a single database. All results are stored in such a way that implementations can directly be compared to each other (using various evaluation measures), and parameter settings are stored so that the impact of individual parameters can be tracked. Moreover, for all data sets, it calculates meta-data about the features and the data distribution\cite{Peng2002}, and for all implementations, meta-data is stored about their (hyper)parameters and properties such as what input data they can handle, what tasks they can solve and, if possible, advanced properties such bias-variance profiles.

Finally, the OpenML website offers various search functionalities. data sets, algorithms and implementations can be found through simple keyword searches, linked to all results and meta-data. Runs can be aggregated to directly compare many implementations over many data sets (e.g. for benchmarking). Furthermore, the database can be queried directly through an SQL editor, or through pre-defined advanced queries.\footnote{See the Advanced tab on \texttt{http://openml.org/search}.} The results of such queries are displayed as data tables, scatterplots or line plots, which can be downloaded directly.

\section{mldata}
mldata ({\footnotesize \texttt{http://mldata.org}}) is a community-based website for the exchange of machine learning data sets. Data sets can either be raw data files or collections of files, or use one of the supported file formats like HDF5 or ARFF in which case mldata looks at meta data contained in the files to display more information. Similar to OpenML, mldata can define learning tasks based on data sets, where mldata currently focuses on supervised learning data. Learning tasks identify which features are used for input and output and also which score is used to evaluate the functions. mldata also allows to create learning challenges by grouping learning tasks together, and lets users submit results in the form of predicted labels which are then automatically evaluated.

mldata.org supports four kinds of information: raw data sets, learning tasks, learning methods, and challenges. A raw data set is just some data, while the learning task also specifies the input and output variables and the cost function used in evaluation. A learning method is the description of a full learning workflow, including feature extraction and learner. One can upload predicted labels for a data set and a task to create a solution entry which automatically evaluates the error on the predicted labels. Finally, a number of learning tasks can be grouped to create a challenge.

Most of this data is text. mldata defines a general file exchange format for supervised learning based on HDF5, a structured compressed file format. It is similar to an archive of files but has additional structure on the level of the files, such that users can directly store and access matrices, or numerical arrays. Using this specified file format is not mandatory, but using it unlocks a number of additional features like a summary of the data set, and automatic conversion into a number of other formats.

Currently, OpenML is being integrated with mldata, so that data sets and learning methods can be shared between both platforms.

\section{Related work}
There also exist platforms aimed at providing reproducible benchmarks. DELVE ({\footnotesize \texttt{http://www.cs.utoronto.ca/\~{}delve}}) was the first, but is currently in abeyance. MLComp ({\footnotesize \texttt{http://mlcomp.org}}) allows users to upload their algorithms and evaluate them on known data sets (or vice versa) on MLComp servers. RunMyCode ({\footnotesize \texttt{http://runmycode.org}}) allows researchers to create \emph{companion websites} for publications by uploading code and building an interface. Users can then fill in all inputs online and get the result of the algorithm.

Compared to these systems, OpenML and mldata allow users more flexibility in running experiments: new tasks can be introduced for novel types of experiments and experiments can be run in any environment. OpenML also offers clean integration in data mining platforms that researchers already use in daily research, and closer data integration so that researchers can reuse results in many ways beyond direct benchmark comparisons, such as meta-learning studies~\cite{Vanschoren2012}.

\subsubsection*{Acknowledgments}

This work is supported by grant 600.065.120.12N150 from the Dutch Fund for Scientific Research (NWO), and by the IST Programme of the European Community, under the PASCAL2 Network of Excellence, IST-2007-216886.

%
%
%

\end{document}